\documentclass[10pt,twocolumn,letterpaper]{article}

\usepackage{cvpr} 
\usepackage{multirow}
\usepackage{colortbl}
\usepackage{pifont}
\usepackage[accsupp]{axessibility}
\newcommand{\tb}[1]{\textbf{#1}}

\definecolor{MyGray}{rgb}{0.85, 0.85, 0.85}
\definecolor{MyGreen}{HTML}{00B050}
\definecolor{MyBlue}{HTML}{058CFA}
\definecolor{cvprblue}{rgb}{0.21,0.49,0.74}
\usepackage[pagebackref,breaklinks,colorlinks,citecolor=cvprblue,linkcolor=red]{hyperref}
\def\paperID{2405} 
\def\confName{CVPR}
\def\confYear{2025}

\title{GuardSplat: Efficient and Robust Watermarking for 3D Gaussian Splatting}

\author{Zixuan Chen$^{1}$\quad Guangcong Wang$^{2}$\quad Jiahao Zhu$^{1}$\quad Jianhuang Lai$^{1,3,4,5}$\quad Xiaohua Xie$^{1,3,4,5}$\thanks{Corresponding author.}\\
$^1$School of Computer Science and Engineering, Sun Yat-sen University, Guangzhou, China\\
$^2$School of Computing and Information Technology, Great Bay University, Dongguan, China\\
$^3$Guangdong Province Key Laboratory of Information Security Technology, China\\
$^4$Key Laboratory of Machine Intelligence and Advanced Computing, Ministry of Education, China\\
$^5$Pazhou Lab (Huangpu), Guangzhou, China\\
\footnotesize\ttfamily{\{chenzx3, zhujh59\}@mail2.sysu.edu.cn, wanggc3@gmail.com, \{stsljh, xiexiaoh6\}@mail.sysu.edu.cn}\\
\small{Project Page: \href{https://narcissusex.github.io/GuardSplat}{https://narcissusex.github.io/GuardSplat} \qquad Code: \href{https://github.com/NarcissusEx/GuardSplat}{https://github.com/NarcissusEx/GuardSplat}}
}

\begin{document}
\maketitle

\begin{abstract}
3D Gaussian Splatting (3DGS) has recently created impressive 3D assets for various applications. 
However, considering \tb{security}, \tb{capacity}, \tb{invisibility}, and training \tb{efficiency}, the copyright of 3DGS assets is not well protected as existing watermarking methods are unsuited for its rendering pipeline.
In this paper, we propose \tb{GuardSplat}, an innovative and efficient framework for watermarking 3DGS assets.
Specifically, \tb{1)} We propose a CLIP-guided pipeline for optimizing the message decoder with minimal costs. 
The key objective is to achieve high-accuracy extraction by leveraging CLIP's aligning capability and rich representations, demonstrating exceptional \tb{capacity} and \tb{efficiency}.
\tb{2)} We tailor a Spherical-Harmonic-aware (SH-aware) Message Embedding module for 3DGS, seamlessly embedding messages into the SH features of each 3D Gaussian while preserving the original 3D structure. 
This enables watermarking 3DGS assets with minimal fidelity trade-offs and prevents malicious users from removing the watermarks from the model files, meeting the demands for \tb{invisibility} and \tb{security}.
\tb{3)} We present an Anti-distortion Message Extraction module to improve \tb{robustness} against various distortions.
Experiments demonstrate that \textbf{GuardSplat} outperforms state-of-the-art and achieves fast optimization speed.
\end{abstract}

\vspace{-2em}
\section{Introduction}
3D representation is a cutting-edge technique in computer vision and graphics, playing a vital role in various domains such as film production, game development, virtual reality, and autonomous driving.
One of the most promising approaches in this field is 3D Gaussian Splatting (3DGS) \cite{3dgs}. 3DGS revolutionizes 3D representation techniques by offering high fidelity, rapid optimization capabilities, and real-time rendering speed,
which enables the creation of impressive 3D assets \cite{liang2024luciddreamer,yuan2024gavatar,szymanowicz2024splatter,yu2024mip,guedon2024sugar,yan2024gs} in the real world.
However, the risk of valuable 3DGS assets being stolen by unauthorized users poses significant losses to creators.
This situation raises an urgent question: \textit{How can we design a method tailored for 3DGS to protect copyright?}

\begin{figure*}
\centering
\includegraphics[width=0.92\textwidth]{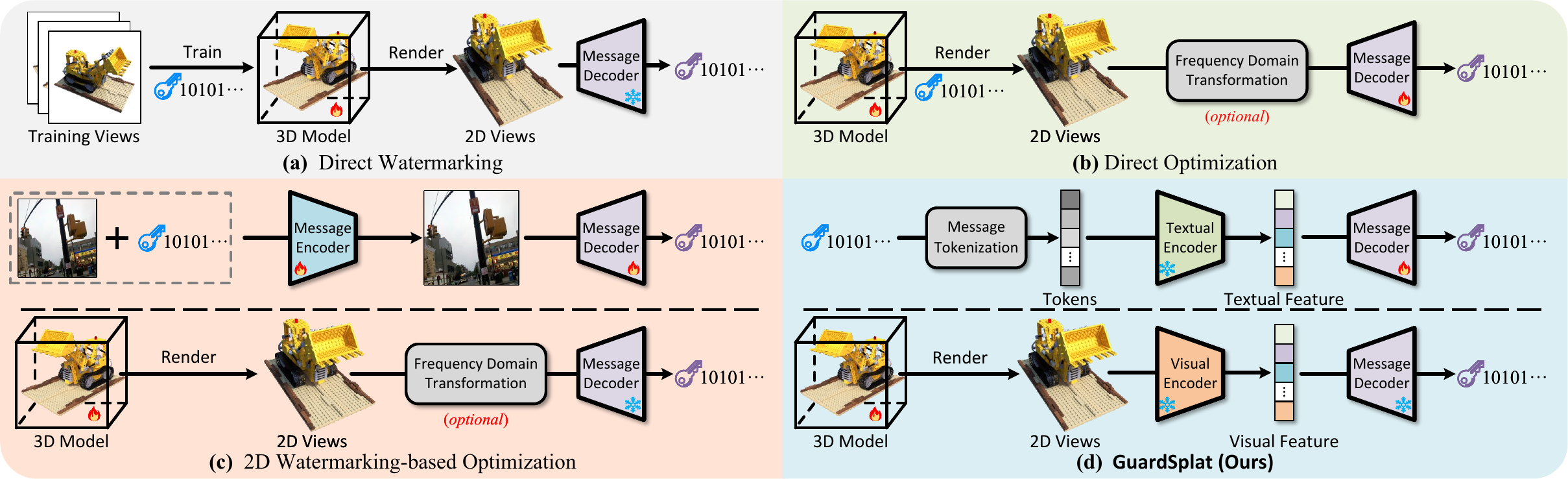}
\caption{\tb{Comparisons of four 3D watermarking frameworks}. They differ in how to embed messages and train message decoders.
\tb{(a)} Directly training 3D models on the watermarked images. \tb{(b)} Simultaneously training a 3D model and a message decoder.
\tb{(c)} Employing the message decoder from a 2D watermarker for optimization.
\tb{(d)} \tb{GuardSplat} first trains a message decoder to extract messages from CLIP \cite{clip} textual features.
This message decoder then can be applied to the CLIP visual features for watermarking 3D models via optimization.}
\label{fig:examples2}
\vspace{-1em}
\end{figure*}

One effective strategy for copyright protection is embedding secret messages into 3DGS assets. 
However, it has several challenging requirements:
\tb{1) Security:} A secure watermark should be difficult to detect and cannot be removed from the model.
\tb{2) Invisibility:} Any watermarked view rendered from a 3D asset should visually maintain consistency with the corresponding view rendered from original models, avoiding disruption of normal use.
\tb{3) Capacity:} Large-capacity messages can be effectively embedded into a 3DGS model and accurately extracted from 2D rendered views.
\tb{4) Efficiency:} Fast optimization speed is essential to meet real-world demands. When considering all of these four important requirements, it is challenging to design a good 3D watermarking method for 3DGS. 

Existing watermarking methods \cite{hidden,copyrnerf,waterf,zhang2024gs} partially meet the four requirements above and have improved a lot for the watermark of 2D or 3D digital assets. However, they are inadequate for the 3DGS framework, which can be categorized into three groups.
\textbf{First}, an intuitive method is directly applying 2D watermarking methods for 2D training or rendered views.
For instance, employing HiDDeN \cite{hidden} to watermark the 2D rendered views cannot protect the original model files.
Besides, directly training 3D models on the watermarked images shown as \Cref{fig:examples2} \tb{(a)} exhibits low bit accuracy in message extraction, as it cannot guarantee novel views contain a consistent watermark. 
\textbf{Second}, some methods directly embed messages into 3D models during optimization (\textit{e.g.,} \cite{copyrnerf,zhang2024gs}, \Cref{fig:examples2} \tb{(b)}). 
Although it guarantees a consistent watermark across novel views, it optimizes a new message decoder per scene during watermarking 3D models, requiring expensive optimization costs. 
\tb{Third}, to avoid per-scene optimization, one might consider a general-purpose message decoder (\textit{e.g.,} \cite{waterf,nerfprotector,gaussianmarker,3d-gsw}, \Cref{fig:examples2} \tb{(c)}), which is pre-trained from a 2D watermarking network.
However, these networks are encoder-decoder models that simultaneously reconstruct the image and extract the message, where the encoder tries to keep the fidelity between the input and output images while the decoder subsequently tries to extract the messages as intact as possible from the output image.
As a result, directly using the decoder to watermark 3D models may yield degraded performance due to the fidelity-capacity trade-off.
Moreover, simultaneously optimizing both encoder and decoder is time-consuming.

\begin{figure}[!t]
\includegraphics[width=0.47\textwidth]{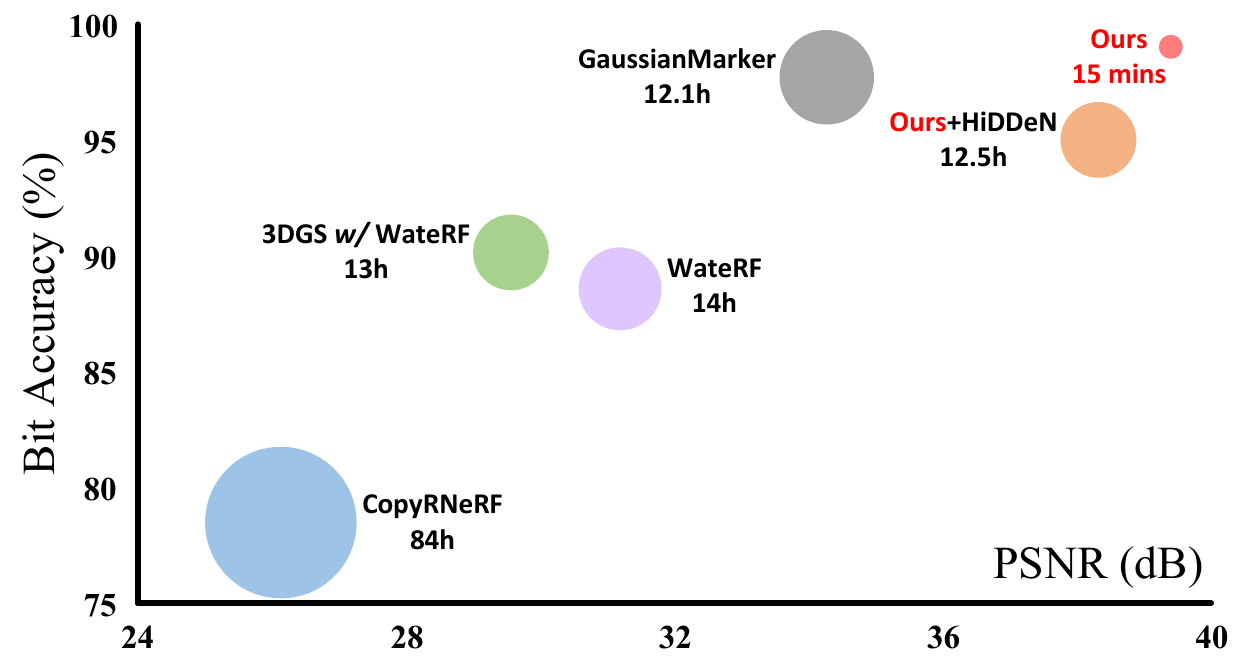}
\caption{
\tb{Performance of state-of-the-art methods} with $N_L=32$ bits on Blender \cite{nerf} and LLFF \cite{llff} datasets.
The radius of circles is proportional to their total training time (decoder optimization + watermarking) evaluated on RTX 3090 GPU.
}
\label{fig:performance}
\vspace{-1em}
\end{figure}

In this paper, we present \textit{GuardSplat}, a novel watermarking framework to protect the copyright of 3DGS assets.
As shown in \Cref{fig:examples2} \tb{(d)}, compared to conventional 2D watermarking methods that optimize both the message encoder and decoder simultaneously, we propose a message decoupling optimization module guided by Contrastive Language-Image Pre-training (CLIP) \cite{clip}, which can only train the decoder for message extraction (top row), significantly reducing the optimization costs.
Thanks to the text-image aligning capability and rich representations of CLIP, this decoder can embed large-capacity messages into 3DGS assets (bottom row) with minimal costs (see Figures \ref{fig:performance} and \ref{fig:time}), demonstrating superior \tb{capacity} and \tb{efficiency}.
Subsequently, we tailor a message embedding module for 3DGS, employing a set of spherical harmonics (SH) offsets to seamlessly embed messages into the SH features of each 3D Gaussian while maintaining the original 3D structure.
This enables watermarking 3DGS assets with minimal trade-offs in fidelity and also prevents malicious users from removing the watermarks from the model files, meeting the demands for \tb{invisibility} and \tb{security}.
We further design an anti-distortion message extraction module, which simulates the randomly distorted views during optimization using the differentiable distortion layer, allowing the watermarked SH features to achieve strong \tb{robustness} against various distortions.
Extensive experiments on Blender \cite{nerf} and LLFF \cite{llff} datasets demonstrate that our \textit{GuardSplat} outperforms the state-of-the-art methods.
\textit{GuardSplat} achieves fast optimization speed, which takes 5 and 10 minutes to train the decoder and watermark a 3DGS asset on a single RTX 3090 GPU, respectively.

Overall, the main contributions of this paper are summarized as follows:
\tb{1)} We present \textit{GuardSplat}, a new watermarking framework to protect the copyright of 3DGS assets. \tb{2)} We propose a CLIP-guided Message Decoupling Optimization module to train the message decoder, achieving superior \tb{capacity} and \tb{efficiency}. We tailor a message embedding method for 3DGS to meet the \tb{invisibility} and \tb{security} demands. We further introduce an anti-distortion message extraction for good \tb{robustness}. \tb{3)} Experiments demonstrate that our \textit{GuardSplat} outperforms state-of-the-art and achieves fast optimization speed for training message decoder and watermarking 3DGS assets.

\section{Related Works}

\textbf{3D Representations.} 
Neural radiance field (NeRF) \cite{nerf} is a compelling solution for 3D representations, which is based on the standard volumetric rendering \cite{rendering} and alpha compositing techniques \cite{alpha}, building the implicit representations using the multi-layer perceptron (MLP). 
Follow-up works adapt NeRF to various domains, such as sparse-view reconstruction \cite{pixnerf,dnerf,Sparsenerf}, acceleration \cite{instngp}, generative modeling \cite{GRAF,Giraffe}, text-to-3D generation \cite{dreamfusion,ProlificDreamer}, anti-aliasing \cite{mip-nerf,mip-nerf-360}, medical image super-resolution \cite{cunerf}, and RGB-D scene synthesis \cite{rgbd-nerf}.
3D Gaussian Splatting (3DGS) \cite{3dgs} has become a mainstream approach for 3D representations with its fast optimization and rendering speed.
Unlike NeRF, 3DGS explicitly represents the scene using a set of 3D Gaussians and renders views through splatting \cite{splatting}.
It has been applied to various scenarios, including text-to-3D generation \cite{liang2024luciddreamer}, avatar generation \cite{shao2024splattingavatar,yuan2024gavatar}, single-view generation \cite{szymanowicz2024splatter,zou2024triplane}, anti-aliasing \cite{yu2024mip}, and SLAM \cite{matsuki2024gaussian,yan2024gs}.

\noindent\textbf{Digital Watermarking.} 
Early studies \cite{barni2001improved, kutter1997digital, raval2003discrete, tao2004robust,dwtdctsvd} proposed to embed the watermarks within frequency domains.
With the advent of deep learning, Zhu \etal \cite{hidden} proposed the first end-to-end deep watermarking framework -- HiDDeN, while the subsequent advances investigate to improve the robustness \cite{luo2020distortion,ssl,cin,WINDinpainting}, and extend the application scenarios \cite{zhang2020udh, StegaStamp, luo2023dvmark}.
Recent methods \cite{fernandez2023stable,wen2024tree} proposed diffusion-based watermarking to protect the contents yielded from diffusion models \cite{ho2020denoising,ddim,Rombach_2022_CVPR,saharia2022photorealistic}.
To protect 3D assets, most methods \cite{ohbuchi2002frequency, praun1999robust, jiahao_zhu_tvcg, zhu2021gaussian} focused on embedding and detecting watermarks within meshes and point clouds.
Yoo \etal \cite{3D-to-2D} provided a novel perspective, which embedded the invisible watermarks into 3D models through differentiable rendering pipelines, allowing the watermarks to be extracted from rendered views.
Inspired by \cite{3D-to-2D}, CopyRNeRF \cite{copyrnerf}, WateRF \cite{waterf}, and NeRFProtector \cite{nerfprotector} aim to insert watermarks into NeRF \cite{nerf}.
Specifically, CopyRNeRF \cite{copyrnerf} replaces the color representations, while WateRF \cite{waterf} and NeRFProtector \cite{nerfprotector} embed the message into model weights via optimization.
Recently, Song \etal \cite{song2024geometry} propose to prevent using Triplane Gaussian Splatting (TGS) \cite{zou2024triplane} for unauthorized 3D reconstruction from copyrighted images. 
Zhang \etal \cite{zhang2024gs} introduced a steganography model for 3DGS that employs secured features to replace SH features, and trains scene and message decoders to extract views and hidden messages, respectively.
Current works \cite{gaussianmarker,3d-gsw} aim to directly embed messages into 3DGS models via a pre-trained 2D watermarking decoder.
However, since 2D watermarking methods have an inherent trade-off between fidelity and capacity, using their decoder for optimization may result in sub-optimal capacity.
Moreover, \cite{gaussianmarker,3d-gsw} may significantly alter the 3D structure during watermarking, leading to low-fidelity results.

\begin{figure*}[!t]
\centering
\includegraphics[width=0.95\textwidth]{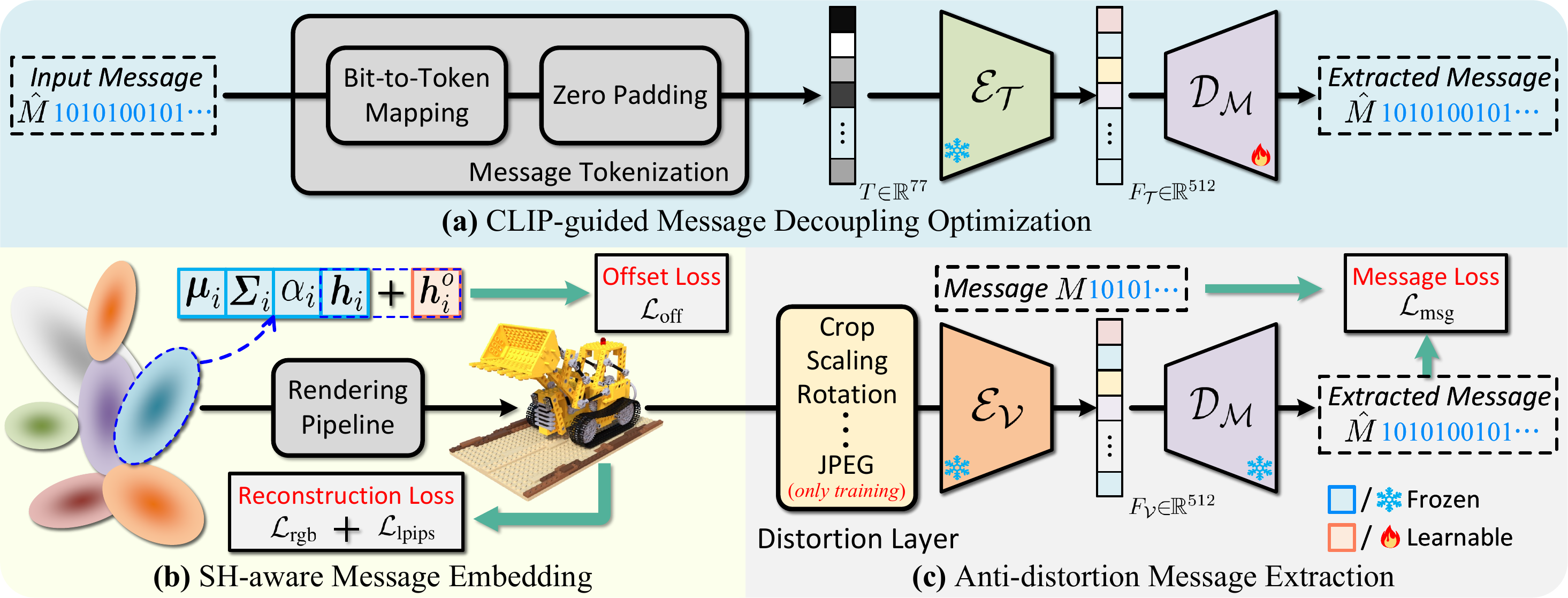}
\caption{\tb{Overview of GuardSplat.} \tb{(a)} Given a binary message $M\in\{0, 1\}^{L}_{i=1}$, we first transform it into CLIP tokens $T$ using the proposed message tokenization.
We then employ CLIP's textual encoder $\mathcal{E_T}$ to map $T$ to the textual feature $F_\mathcal{T}$.
Finally, we feed $F_\mathcal{T}$ into message decoder $\mathcal{D_M}$ to extract the message $\hat{M}\in\{0, 1\}^{L}_{i=1}$ for optimization.
\tb{(b)} For each 3D Gaussian, we freeze all the attributes and build a learnable spherical harmonic (SH) offset $\boldsymbol{h}^o_i$ as the watermarked SH feature, which can be added to the original SH features as $\boldsymbol{h}_i + \boldsymbol{h}^o_i$ to render the watermarked views.
\tb{(c)} We first feed the 2D rendered views to CLIP's visual encoder $\mathcal{E_V}$ to acquire the visual feature $F_{\mathcal{V}}$ and then employ the pre-trained message decoder to extract the message $\hat{M}$.
A differentiable distortion layer is used to simulate various visual distortions during optimization.
$\mathcal{D_M}$ and $\boldsymbol{h}^o_i$ are optimized by \cref{eq:message_loss} and \cref{eq:loss}, respectively.
}
\label{fig:overall}
\vspace{-1em}
\end{figure*}

\section{Preliminary}
\textbf{3D Gaussian Splatting.} 3DGS \cite{3dgs} is a recent groundbreaking method for novel view synthesis.
Given the center position $\boldsymbol{\mu}\in\mathbb{R}^3$ and covariance $\boldsymbol{\varSigma}\in\mathbb{R}^{7}$, a 3D Gaussian at position $\mathbf{x}$ can be queried as follows:
\begin{equation}
\label{eq:gaussian}
\mathcal{G}(\mathbf{x}:\boldsymbol{\mu},\boldsymbol{\varSigma})=\exp(-\frac{1}{2}(\mathbf{x}-\boldsymbol{\mu})^{\top}\boldsymbol{\varSigma}^{-1}(\mathbf{x}-\boldsymbol{\mu})).
\end{equation}
Then, given the projective transformation $\mathbf{P}$, viewing transformation $\mathbf{W}$, and Jacobian $\mathbf{J}$ of the affine approximation of $\mathbf{P}$, the corresponding 2D mean position $\hat{\boldsymbol{\mu}}$ and covariance $\hat{\boldsymbol{\varSigma}}$ of the projected 3D Gaussian can be calculated as:
\begin{equation}
\hat{\boldsymbol{\mu}}=\boldsymbol{P}\boldsymbol{W}\boldsymbol{\mu}, \quad \hat{\boldsymbol{\varSigma}}=\mathbf{J}\mathbf{W} \boldsymbol{\varSigma} \mathbf{W^{\top}}\mathbf{J^{\top}}.
\end{equation}
Let $\hat{C} \in \mathbb{R}^{W\times H\times3}$ denote a $W\times H$ RGB view rendered by 3DGS, the color of each pixel $(x,y)$ can be generated as:
\begin{equation}
\hat{C}_{x,y}\! = \!\!\sum_{i=1}^{N}\!\boldsymbol{c}_{i}\sigma_{i}\!\prod_{j=1}^{i-1}(1-\sigma_{j}),\quad
\sigma_i\!=\!\alpha_i\mathcal{G}((x,y)\!:\!\hat{\boldsymbol{\mu}},\hat{\boldsymbol{\varSigma}}),
\label{eq:3dgs_rendering}
\end{equation}
where $N$ represents the number of Gaussians overlaping the pixel $(x,y)$.
$\boldsymbol{c}_{i} \in \mathbb{R}^{3}$ and ${\alpha}_{i} \in \mathbb{R}^1$ denote the color transformed from $k$-ordered spherical harmonic (SH) coefficients $\boldsymbol{h}_i \in \mathbb{R}^{3\times (k+1)^2}$ and opacity of the $i$-th Gaussian, respectively.
Since the rendering pipeline is differentiable, 3DGS models can be optimized by the loss function as:
\begin{equation}
\mathcal{L}_\text{rgb}=\lambda_\text{ssim}\mathcal{L}_\text{ssim}(\hat{C}, C)+(1-\lambda_\text{ssim})\mathcal{L}_1(\hat{C}, C),
\label{eq:rgb_loss}
\end{equation}
where $C$ is a groundtruth image and $\lambda_\text{ssim}$ is set to 0.2.

\noindent\textbf{Contrastive Language-Image Pre-training.}\label{sec:clip} CLIP \cite{clip} is pre-trained to match images with natural language descriptions on 400 million training image-text pairs collected from the internet.
It consists of two independent encoders: a textual encoder $\mathcal{E_T}$ and a visual encoder $\mathcal{E_V}$, which extract textual features $F_{\mathcal{T}}\in\mathbb{R}^{512}$ and visual features $F_{\mathcal{V}}\in\mathbb{R}^{512}$ from the given batch of images and texts, respectively.
These encoders are trained to learn the aligning capability of text-image pairs by maximizing the similarity between textual and visual features via a contrastive loss.

\section{Method}
In this section, we propose \textit{GuardSplat} to effectively protect the copyright of 3D Gaussian Splatting (3DGS) \cite{3dgs} assets.
The overview of the proposed method is depicted in \Cref{fig:overall}.
Specifically, we first propose a message decoupling optimization module guided by Contrastive Language-Image Pretraining (CLIP) \cite{clip} to train the message decoder $\mathcal{D_M}$.
By analyzing the 3DGS rendering pipeline, we then present a spherical-harmonic-aware (SH-aware) message embedding module to integrate the watermarked SH features for a pre-trained 3DGS model.
Furthermore, we design a strategy for anti-distortion message extraction.
As a result, \textit{GuardSplat} can not only embed large-capacity messages with minimal optimization costs but also achieves superior fidelity and robustness, suggesting that \textit{GuardSplat} can protect the copyright of 3DGS models without the affection of normal use.

\subsection{Message Decoupling Optimization}\label{sec:pre-training}
As discussed in recent works \cite{3D-to-2D,copyrnerf,waterf}, one of the representative approaches to watermark 3D assets is embedding the messages into a 3D representation model via optimization.
However, these methods encounter limitations in efficiency.
Specifically, one group of methods aims to directly embed messages into 3D models for optimization, as shown in \Cref{fig:examples2} \tb{(b)}. They optimize a message decoder per scene, which is time-consuming.
To learn a general-purpose message decoder, the other group of methods first trains an encoder and decoder such that it can reconstruct the image and extract the message given an image and message as input. The message decoder is then used for 3D watermarking as shown in \Cref{fig:examples2} \tb{(c)}.
Since conventional methods optimize both the message encoder and decoder simultaneously, it also takes much time for optimization.

To address this issue, we propose a Message Decoupling Optimization module guided by CLIP \cite{clip} that optimizes a general-purpose message decoder and a 3DGS model, as shown in \Cref{fig:examples2} \tb{(d)}. The key to the success of this module is that the Contrastive Language-Image Pre-training (CLIP) \cite{clip} builds a bridge between the texts and images. As discussed in \Cref{sec:clip}, CLIP consists of a good textual encoder and a good visual encoder that is trained on a dataset of 400 million text-image pairs, providing rich text and image representations. As shown in \Cref{fig:overall} \tb{(a)}, given a binary message $M\in\{0, 1\}^{L}_{i=1}$, we first transform it into CLIP tokens $T$ using the proposed bit-to-token mapping as:
\begin{equation}
T = \{t_S\}\cup\left\{\bigcup_{i=1}^L\varPhi(M_i, i)\right\}\cup\{t_E\},
\end{equation}
where $t_S=49406$ and $t_E=49407$ denote the start and ending points of the CLIP text token, respectively.
$\varPhi(\cdot, \cdot)$ is a function that uniformly maps the $i$-th bit to an integer number within the range $[1, 49405]$.
To match the format of CLIP tokens, $T$ is then zero-padded to a size of $77$.
Finally, we feed the tokens into CLIP textual encoder $\mathcal{E_T}$ and employ a message decoder $\mathcal{D_M}$ built by multi-layer perception (MLP) with 3 fully-connected (FC) layers to extract the messages $\hat{M}$ from the output textual features $F_\mathcal{T}\in\mathbb{R}^{512}$ as:
\begin{equation}
\hat{M}=\mathcal{D_M}(\mathcal{E_V}(T)).
\end{equation}
$\mathcal{D_M}$ can be optimized by minimizing the message loss as:
\begin{equation}
\mathcal{L}_\text{msg} = -\sum^L_{i=1}M_i\log\hat{M}_i+(1-M_i)\log(1-\hat{M}_i).
\label{eq:message_loss}
\end{equation}
As a result, the message decoder can be directly optimized without being constrained by invisibility, achieving superior capacity and training efficiency.
Furthermore, CLIP's rich representation also enables it to achieve better performance with minimal optimization costs.

\subsection{SH-aware Message Embedding}
Unlike neural radiance fields (NeRF) \cite{nerf}, 3D Gaussian Splatting (3DGS) \cite{3dgs} explicitly represents the scene through a set of 3D Gaussian as \cref{eq:gaussian} and renders the views using \cref{eq:3dgs_rendering}.
Given the $i$-th 3D Gaussian, it consists of 4 attributes: center position $\boldsymbol{\mu}_i$, covariance $\boldsymbol{\varSigma}_i$, opacity $\boldsymbol{\alpha}_i$, and spherical harmonic (SH) feature $\boldsymbol{h}_i$, where the former three attributes denote the 3D structure and while the latter is related to the color representation.
An intuitive solution to watermark 3DGS assets is directly updating all the attributes of 3D Gaussians during optimization.
However, it may significantly alter the 3D structure (i.e., $\boldsymbol{\mu}_i$, $\boldsymbol{\varSigma}_i$, $\boldsymbol{\alpha}_i$), which leads to sub-optimal fidelity while concealing large-capacity messages (see ``Offset$_{\text{all}}$'' in \Cref{table:diff_variants}). 

Based on the above observations, we argue that it is required to maintain the original 3D structure during optimization.
To achieve this, we propose an SH-aware Message Embedding module, a simple yet efficient approach tailored for 3DGS to watermark pre-trained models with minimal losses in fidelity.
As shown in \Cref{fig:overall} \tb{(b)}, for each 3D Gaussian, we freeze all the attributes and create a learnable SH offset $\boldsymbol{h}^o_i\in\mathbb{R}^{48}$ for watermarking. The reason behind this is the fact that SH parameters represent view-dependent effects like glossy or specular highlights, which only exist in a few regions of a scene. Thus, embedding the secret message into SH features with minimal constraints can preserve the fidelity of the 3D asset. 
Specifically, we first add each SH offset $\boldsymbol{h}^o_i$ to the corresponding SH coefficient $\boldsymbol{h}_i$ as the watermarked SH feature, and then feed it into the 3DGS rasterization to render the watermarked views.
To further alleviate the fidelity decline brought by excessive offset, we employ an offset loss to constrain its magnitude:
\begin{equation}
\mathcal{L}_\text{off}=-\frac{1}{N}\sum^N_{i=1}\|\boldsymbol{h}^o_i\|^2_2,
\end{equation}
where $N$ denotes the number of 3D Gaussians.
As a result, the secret message can be seamlessly embedded into the SH offset of each 3D Gaussian in optimization, which not only maintains the original 3D structure but also prevents malicious users from removing the watermarks from the model files, achieving superior invisibility and security.

\begin{figure}[!t]
\centering
\includegraphics[width=0.47\textwidth]{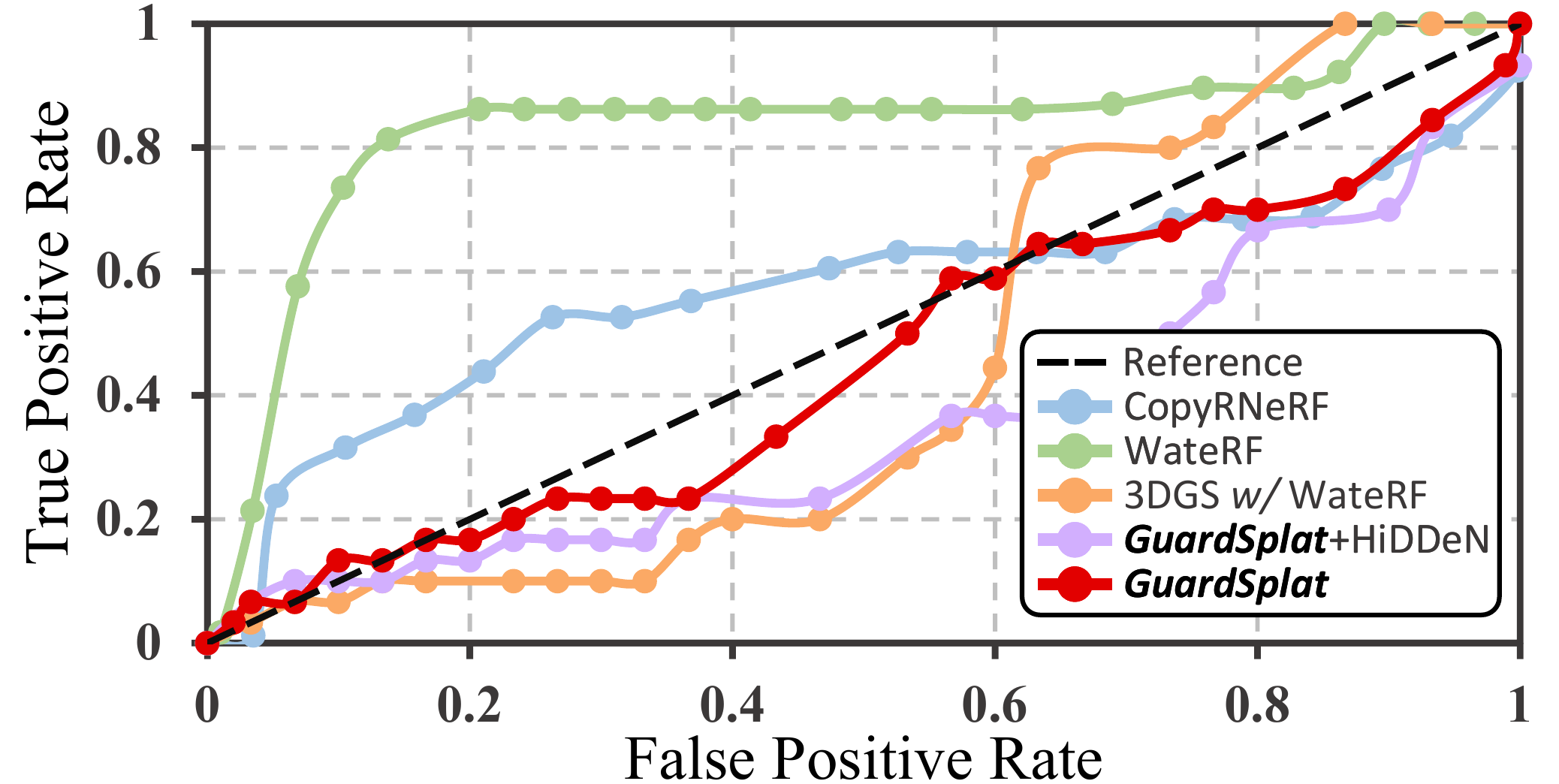}
\caption{\tb{ROC curves} produced by varying thresholds in StegExpose \cite{stegexpose} on different methods.
The closer the curve is to the ``Reference'', the more effective the method is regarding security.}
\label{fig:security}
\vspace{-1em}
\end{figure}

\begin{table*}[!t]
\caption{\tb{Comparisons of the start-of-the-art methods} on Blender \cite{nerf} and LLFF \cite{llff} datasets for bit accuracy and reconstruction qualities \textit{w.r.t} various message lengths.
\tb{Bold} text indicates the best performance in this table.
}
\setlength{\tabcolsep}{1.2mm}
\centering
\footnotesize
\label{table:allresults}
\resizebox{0.95\textwidth}{!}{
\begin{tabular}{lcccccccccccccc}
\toprule  
\multirow{2}{*}{Methods}  &\multicolumn{4}{c}{16 bits}  &&\multicolumn{4}{c}{32 bits}  &&\multicolumn{4}{c}{48 bits}  \\\cmidrule{2-5}\cmidrule{7-10}\cmidrule{12-15}
&Bit Acc  &PSNR   &SSIM    &LPIPS   &&Bit Acc  &PSNR   &SSIM    &LPIPS   &&Bit Acc  &PSNR   &SSIM    &LPIPS   \\\midrule
\multicolumn{15}{l}{\scriptsize\textit{NeRF-based Watermarking Methods}}\\
CopyRNeRF \cite{copyrnerf}              &91.16   &26.29   &0.9100   &0.0380   &&78.08   &26.13   &0.8960   &0.0410   &&60.06   &27.56   &0.8950   &0.0660   \\
WateRF \cite{waterf}                 &95.67   &32.79   &0.9480   &0.0330   &&88.58   &31.19   &0.9360   &0.0400   &&85.82   &30.86   &0.9300   &0.0400   \\\midrule
\multicolumn{15}{l}{\scriptsize\textit{3DGS Built on Watermarked Images}}\\
3DGS \cite{3dgs} + CIN \cite{cin}          &56.73   &31.95   &0.9194   &0.1027   &&53.13   &31.74   &0.9279   &0.0944   &&55.78   &30.25   &0.9139   &0.1120   \\
3DGS \cite{3dgs} + SSL \cite{ssl}          &58.94   &36.51   &0.9737   &0.0094   &&61.85   &35.24   &0.9706   &0.0179   &&58.79   &35.88   &0.9710   &0.0123   \\
3DGS \cite{3dgs} + HiDDeN \cite{hidden}       &63.07   &31.59   &0.9790   &0.0171   &&52.46   &34.51   &0.9682   &0.0209   &&53.08   &33.42   &0.9687   &0.0299   \\
3DGS \cite{3dgs} + DwtDctSvd \cite{dwtdctsvd}    &55.44   &34.78   &0.9582   &0.0399   &&53.15   &32.32   &0.9477   &0.0547   &&51.83   &31.09   &0.9302   &0.0704   \\
3DGS \cite{3dgs} + StegaStamp \cite{StegaStamp}   &79.72   &35.52   &0.9697   &0.0181   &&82.36   &34.04   &0.9601   &0.0265   &&83.97   &32.54   &0.9523   &0.0406   \\\midrule
\multicolumn{15}{l}{\scriptsize\textit{3DGS optimized by altering all attributes} ($\text{Offset}_\text{all}$)}\\
GaussianMarker \cite{gaussianmarker}         &99.36   &34.42   &0.9822   &0.0124   &&98.85   &33.98   &0.9788   &0.0163   &&98.25   &32.12   &0.9723   &0.0234   \\
3DGS \cite{3dgs} \textit{w/} WateRF \cite{waterf}  &92.89   &31.01   &0.9678   &0.0475   &&90.15   &29.56   &0.9611   &0.0492   &&87.30   &29.13   &0.9562   &0.0534   \\\midrule
\multicolumn{15}{l}{\scriptsize\textit{GuardSplat (Ours) + 2D Watermarking Decoders}}\\
\tb{GuardSplat} (Ours) + CIN \cite{cin}       &95.75   &37.88   &0.9762   &0.0092   &&93.35   &37.42   &0.9726   &0.0109   &&92.77   &37.10   &0.9689   &0.0124   \\
\tb{GuardSplat} (Ours) + SSL \cite{ssl}       &99.50   &40.92   &0.9935   &0.0020   &&98.60   &38.95   &0.9920   &0.0028   &&98.14   &38.51   &0.9909   &0.0030   \\
\tb{GuardSplat} (Ours) + HiDDeN \cite{hidden}    &98.75   &40.48   &0.9909   &0.0025   &&95.58   &38.32   &0.9897   &0.0025   &&93.29   &38.56   &0.9886   &0.0032   \\
\tb{GuardSplat} (Ours) + StegaStamp \cite{StegaStamp}&99.00   &38.55   &0.9903   &0.0035   &&98.28   &38.63   &0.9914   &0.0030   &&97.23   &38.27   &0.9892   &0.0037   \\
\rowcolor{MyGray}   
\tb{GuardSplat} (Ours)                &\tb{99.64}&\tb{41.55}&\tb{0.9957}&\tb{0.0017}&&\tb{99.04}&\tb{39.40}&\tb{0.9939}&\tb{0.0022}&&\tb{98.29}&\tb{38.90}&\tb{0.9923}&\tb{0.0028}\\\bottomrule
\end{tabular}
}
\vspace{-1em}
\end{table*}

\subsection{Anti-distortion Message Extraction} 
Given CLIP's visual encoder $\mathcal{E_V}$ and the message decoder $\mathcal{D_M}$ pre-trained in \Cref{sec:pre-training}, we can directly extract the hidden messages $\hat{M}$ from the watermarked views $\hat{C}$ based on CLIP's aligning capability by: 
\begin{equation}
\hat{M}=\mathcal{D_M}(\mathcal{E_V}(\hat{C}),
\label{eq:extraction}
\end{equation}
Thanks to the generalization capability of CLIP, the watermarked color features have strong robustness against the visual distortions of Gaussian blur, and Gaussian noise.
However, it struggles to deal with other visual distortions such as rotation and JPEG compression.
To robustly extract messages from the watermarked views across various types of distortions, we further propose an anti-distortion message extraction module.
In \Cref{fig:overall} \tb{(c)}, we employ a differentiable distortion layer to randomly simulate some visual distortions on the rendered views during optimization. 
These distortions include cropping, scaling, rotation, JPEG compression, and brightness jittering, enabling the watermarked SH features to learn the anti-distortion ability against most visual distortions via optimization.
The distortion layer is only used during training.
After optimization, \textit{GuardSplat} can deal with most visual distortions, achieving superior extraction accuracy under challenging conditions.

\subsection{Full Objective}
For optimization, we first freeze CLIP's visual and textual encoders and train the message decoder by minimizing the message loss $\mathcal{L}_\text{msg}$ in \cref{eq:message_loss} between the input and extracted messages.
After pre-training the CLIP-guided message decoder, we then employ it to watermark the pre-trained 3DGS models.
We freeze the message decoder and utilize it to extract the message from the rendered views, and the secret message can be embedded into 3DGS models by minimizing the following loss:
\begin{equation}
\mathcal{L} = \lambda_\text{recon}(\mathcal{L}_\text{rgb} + \mathcal{L}_\text{lpips}) + \lambda_\text{msg}\mathcal{L}_\text{msg} + \lambda_\text{off}\mathcal{L}_\text{off},
\label{eq:loss}
\end{equation}
where $\lambda_\text{recon}$, $\lambda_\text{msg}$, and $\lambda_\text{off}$ are the hyper-parameters to balance the corresponding terms.
Since $\mathcal{L}_\text{rgb}$ and $\mathcal{L}_\text{lpips}$ separately denote the RGB loss in \cref{eq:rgb_loss} and LPIPS loss \cite{lpips_loss}, optimizing the reconstruction term $\mathcal{L}_\text{recon}=\mathcal{L}_\text{rgb} + \mathcal{L}_\text{lpips}$ can improve the visual similarity between the watermarked and original views.
For extraction, given a 3DGS model watermarked by our \textit{GuardSplat}, we can render the views from arbitrary viewpoints using the official rendering pipeline, while the hidden messages can be directly extracted from the rendered views by \cref{eq:extraction} for copyright identification.

\begin{table*}[!t]
\caption{\tb{Comparisons of the start-of-the-art methods} on Blender \cite{nerf} and LLFF \cite{llff} datasets for bit accuracy \textit{w.r.t} various distortion types.
We show the results on 16-bit messages.
\tb{Bold} text indicates the best performance in this table.
}
\centering
\footnotesize
\setlength{\tabcolsep}{0.95mm}
\label{table:robustness}
\resizebox{0.95\textwidth}{!}{
\begin{tabular}{lcccccccccc}
\toprule
\multirow{2}{*}{Methods}               &\multirow{2}{*}{None} &Noise   &Rotation   &Scaling   &Blur      &Crop   &Brightness  &JPEG     &VAE Attack \cite{vae_attack}&\multirow{2}{*}{Combined}\\
&           &($\mu$=0.1)&($\pm$$\pi$/6)&($\leq$25\%)&($\sigma$=$0.1$)&(40\%)  &(0.5$\sim$1.5)&(10\% quality)&Bmshj2018          &             \\\midrule
CopyRNeRF \cite{copyrnerf}              &91.16         &90.04   &88.13     &89.33    &90.06      &--    &--      &--      &--             &--            \\
WateRF \cite{waterf}                 &95.67         &95.36   &93.13     &93.29    &95.25      &95.40   &90.91     &86.99     &51.73            &84.12          \\
3DGS \cite{3dgs} \textit{w/} WateRF \cite{waterf}  &92.89         &87.35   &88.28     &90.33    &91.92      &89.07   &88.71     &88.49     &55.48            &86.37          \\
GaussianMarker \cite{gaussianmarker}         &99.36         &99.13   &70.84     &97.89    &94.40      &98.52   &95.78     &86.22     &52.00            &83.49          \\
\tb{GuardSplat} (Ours) + CIN \cite{cin}       &95.75         &94.87   &90.89     &94.50    &95.16      &93.82   &93.97     &88.61     &49.25            &84.03          \\
\tb{GuardSplat} (Ours) + SSL \cite{ssl}       &99.50         &99.57   &86.78     &84.53    &98.79      &77.54   &94.31     &92.99     &47.42            &74.85          \\
\tb{GuardSplat} (Ours) + HiDDeN \cite{hidden}    &98.75         &96.63   &90.02     &95.93    &94.87      &97.25   &94.97     &90.04     &53.14            &88.70          \\
\tb{GuardSplat} (Ours) + StegaStamp \cite{StegaStamp}&99.00         &98.38   &53.21     &95.17    &98.17      &51.34   &95.48     &88.81     &80.12            &64.75          \\
\rowcolor{MyGray} 
\tb{GuardSplat} (Ours)                &\tb{99.64}      &\tb{99.60} &\tb{94.56}  &\tb{98.75} &\tb{99.27}   &\tb{98.71}&\tb{97.46}  &\tb{94.70}  &\tb{82.35}         &\tb{93.38}        \\\bottomrule
\end{tabular}
}
\vspace{-1em}
\end{table*}

\section{Experiments}
\noindent\tb{Datasets.}
Following the settings in \cite{copyrnerf,waterf}, we choose two commonly-used datasets: the Blender \cite{nerf} and LLFF \cite{llff} datasets, for evaluation.
Specifically, the Blender dataset consists of 8 synthetic bounded scenes, while the LLFF dataset consists of handheld forward-facing captures of 8 real scenes.
For each scene, we employ 200 test views to evaluate the visual quality and bit accuracy.
We report average values across all testing views in our experiments.

\noindent\tb{Baselines.}
We compare \textit{GuardSplat} against six baselines to ensure a fair comparison: \textbf{1)} CopyRNeRF \cite{copyrnerf}, \textbf{2)} WateRF \cite{waterf}, \textbf{3)} GaussianMarker \cite{gaussianmarker}, \textbf{4)} 3DGS \textit{w/} WateRF, \textbf{5)} 3DGS trained on watermarked images (\textit{i.e.,} 3DGS + 2D watermarking methods), and \textbf{6)} \textit{GuardSplat} optimized by other pre-trained 2D watermarking decoders, (\textit{i.e.,} \textit{GuardSplat} + 2D watermarking decoders).
Specifically, CopyRNeRF and WateRF are NeRF-based \cite{nerf} watermarking methods, while GaussianMarker is a 3DGS watermarking approach. 
3DGS \textit{w/} WateRF refers to applying WateRF to 3DGS models.
The 2D watermarking methods include DwtDctSvd \cite{dwtdctsvd}, StegaStamp \cite{StegaStamp}, SSL \cite{ssl}, and CIN \cite{cin}.

\begin{figure*}
\centering
\includegraphics[width=\textwidth]{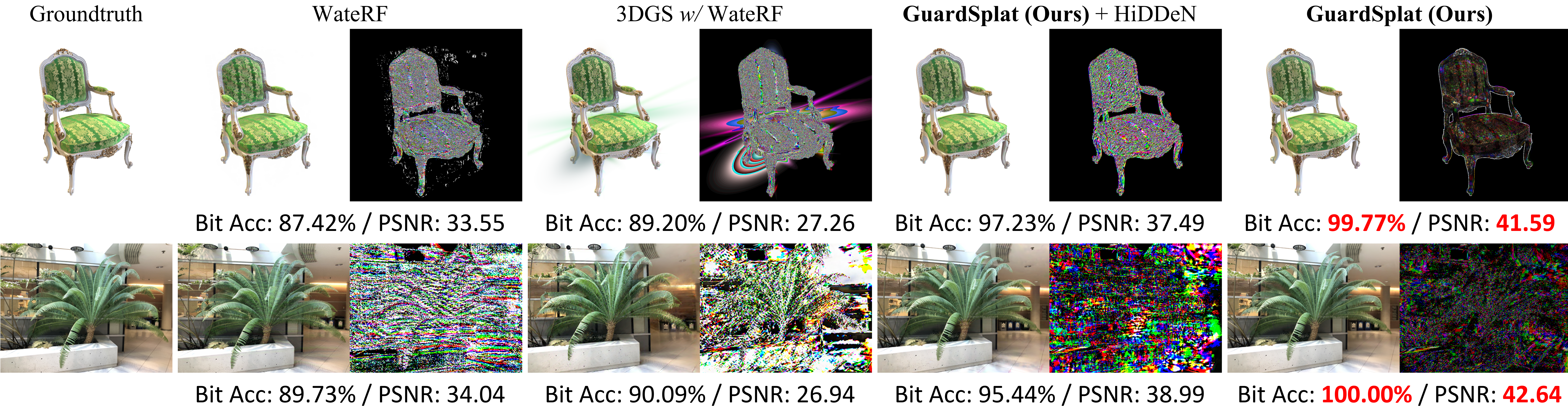}
\caption{\tb{Visual comparisons} with $N_L=32$ bits. Heatmaps show the differences ($\times10$) between the watermarked and original views.}
\label{fig:visual}
\vspace{-1em}
\end{figure*}

\noindent\tb{Implementation Details.}
\textit{GuardSplat} is implemented on the top of a Pytorch \cite{Pytorch} implementation of 3DGS \cite{3dgs}\footnote{https://github.com/graphdeco-inria/gaussian-splatting}.
Our experiments run on a single RTX 3090 GPU.
We first train a 3DGS model on a given scene with multi-view photos and then embed the message into that pre-trained 3DGS model.
The size of learnable SH offsets is equal to that of the SH coefficients in the pre-trained models.

\noindent\textit{For training the message decoder}, we employ Adam \cite{adam} as the optimizer with a weight decay of $10^{-6}$ and a batch size of $64$.
The maximum epoch is set to 100, and the learning rate is set to $5$$\times10^{-3}$ as default.
Given the message length $N_L$, we randomly select $\mathtt{min}(2^{N_L}, N_K)$ samples from a total of $2^{N_L}$ messages as training and test data, where $N_K$=$2048$.
It only takes 5 minutes for optimization.
Thanks to CLIP's rich representation, our decoder achieves excellent performance with only 3 FC layers, detailed architecture is provided in Supp. \ref{sec:Decoder}.

\noindent\textit{For watermarking 3DGS models}, we employ Adam \cite{adam} as the optimizer with a weight decay of $10^{-6}$ and a batch size of $16$.
The maximum epoch is set to 100, and the learning rate of the SH offsets is set to $5$$\times$$10^{-3}$.
The hyper-parameters in \cref{eq:loss} are set as $\lambda_\text{recon}$=$1$, $\lambda_\text{msg}$=$0.03$, and $\lambda_\text{off}$=$10$, respectively.
It takes 10 minutes for watermarking.

\noindent\textit{For Distortion Layer}, since the proposed CLIP-guided message decoder can deal with Gaussian Noise and Gaussian Blur, we only consider the visual distortions of cropping, scaling, rotation, brightness jittering, JPEG compression, and VAE attacks.
The former four visual distortions are differentiably re-implemented by Pytorch \cite{Pytorch} built-in functions, while the differentiable JPEG comparison is built by \cite{diff_jpeg}.
During training, we employ the differentiable distortion layer to simulate various visual distortions for optimization.
In the test, the rendered views are distorted using the visual distortions built by OpenCV for evaluation.

\begin{figure*}[!t]
\centering
\includegraphics[width=\textwidth]{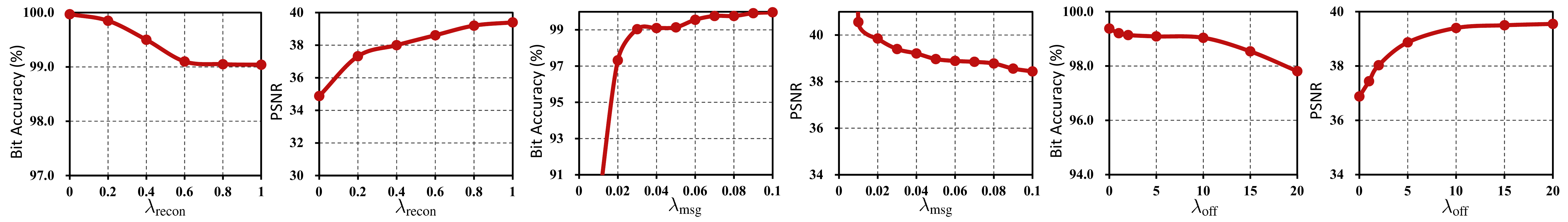}
\caption{\tb{Performance across different hyper-parameter values} with $N_L=32$ bits on the Blender \cite{nerf} and LLFF \cite{llff} datasets.}
\label{fig:sensitivity}
\vspace{-1em}
\end{figure*}

\noindent\tb{Evaluation Metrics.}
We follow the standard of digital watermarking to evaluate \textit{GuardSplat} and the baselines in five aspects:
\textbf{1) Capacity:} We evaluate the bit accuracy across various message lengths $N_L\in\{16,32,48\}$ on the 2D rendered views.
\textbf{2) Invisibility:} We evaluate the visual similarity of views rendered from the watermarked and original models using Peak Signal-to-Noise Ratio (PSNR), Structured Similarity Index (SSIM) \cite{ssim}, and Learned Perceptual Image Patch Similarity (LPIPS) \cite{lpips}.
\textbf{3) Robustness:} We investigate the extraction accuracy with $N_L=16$ bits across various visual distortions, including Gaussian Noise of $\mu$=0 and $\sigma$=0.1, Random Rotation of angles within $[-\frac{\pi}{6}, +\frac{\pi}{6}]$, Random Scaling of ratios within $[0.75, 1.25]$, Gaussian Blur of \textit{kernel\_size}=3 and $\sigma$=0.1, 40\% Center Crop, Brightness Jittoring of ratios within $[0.5, 1.5]$, JPEG compression of 10\% image quality, VAE attack \cite{vae_attack} based on Bmshj2018, and the combination of Crop, Brightness, and JPEG.
\textbf{4) Security:} We perform StegExpose \cite{stegexpose}, an LSB steganography detection, on the rendered views.
The detection set is built by mixing the watermarked and original views with equal proportions. 
\textbf{5) Efficiency:} We analyze the relationship between bit accuracy and training time.

\subsection{Experimental Results}\label{sec:comp}
\noindent\tb{Security.}
We claim that our \textit{GuardSplat} is secure.
Since it adaptively embeds messages by slightly perturbing the SH features of every 3D Gaussian, it is difficult to remove the watermark from the model file.
To further verify the security, we employ StegExpose \cite{stegexpose} to detect steganographic content in rendered views.
\Cref{fig:security} depicts the detection results on our \textit{GuardSplat} and the baselines.
As shown, our \textit{GuardSplat} achieves superior security to the competitors.

\noindent\tb{Capacity \& Invisibility.}
We report the capacity and invisibility across various message lengths $N_L$$\in$$\{16,32,48\}$ on the Blender \cite{nerf} and LLFF \cite{llff} datasets in terms of bit accuracy, PSNR, SSIM, and LPIPS in \Cref{table:allresults}.
As demonstrated, our \textit{GuardSplat} surpasses all the competitors with a consistently superior performance \textit{w.r.t} various message lengths, achieving a significant improvement in extraction accuracy and invisibility.
Moreover, \textit{GuardSplat} achieves higher bit accuracy than \textit{GuardSplat} + 2D watermarking decoders with $N_L\geq32$ bits, indicating that our CLIP-guided message decoder is better at handling large-capacity messages.
Besides, \textit{GuardSplat} + HiDDeN outperforms 3DGS \textit{w/} WateRF as they share the same decoder, proving the superiority of our SH-aware Message embedding module.

\noindent\tb{Robustness.}
We report bit accuracy across various visual distortions with $N_L=16$ bits in \Cref{table:robustness}.
Visually, our \textit{GuardSplat} outperforms all the baselines across various distortions, demonstrating that our Anti-distortion Message Extraction module enables the watermarked 3DGS models to learn robust SH features during optimization.

\noindent\tb{Efficiency.}
As shown in \Cref{fig:performance}, the efficiency of existing advances is unsatisfactory.
Specifically, training a CopyRNeRF requires 84 hours, while WateRF, and 3DGS \textit{w/} WateRF separately spend 14 and 13 hours for optimization as it takes 12 hours to train the HiDDeN.
Compared to these methods, our \textit{GuardSplat} achieves much higher efficiency, which only takes 5 and 10 minutes to train the message decoder and watermark a 3DGS asset, respectively.
We also provide the comparison of watermarking speed in Supp. \ref{sec:Decoder}.

\begin{table}[!t]
\caption{\tb{Various embedding methods} on Blender \cite{nerf} and LLFF \cite{llff} datasets with $N_L$=$32$ bits. 
\tb{Bold} text denotes the best score.}
\label{table:diff_variants}
\centering
\setlength{\tabcolsep}{1.7mm}
\resizebox{0.88\linewidth}{!}{
\begin{tabular}{lcccc}
\toprule
& Bit Acc & PSNR & SSIM  & LPIPS \\\midrule
Offest$_{\text{all}}$  & 98.79  & 36.56 & 0.9804 & 0.0123 \\
Offset$_{\text{dc}}$   & 74.59  & 36.98 & 0.9828 & 0.0146\\
Offest$_{\text{rest}}$  & 98.25  & 38.70 & 0.9892 & 0.0077\\
\rowcolor{MyGray}
\tb{SH-aware} (Ours)  & 99.04  & 39.40 & 0.9939 & 0.0022 \\\bottomrule
\end{tabular}}
\vspace{-1em}
\end{table}

\noindent\tb{Visual Comparison.}
We visually compare our \textit{GuardSplat} against baselines with $N_L=32$ bits in \Cref{fig:visual}.
As shown, our results present superior reconstruction quality and bit accuracy to the competitors.
Moreover, the fidelity of 3DGS \textit{w/} WateRF is much lower than \textit{GuardSplat} + HiDDeN, which demonstrates that altering all the attributes during optimization may lead to a significant decline in image quality.

\subsection{Ablation Study \& Sensitivity Analysis}\label{sec:abla}
We first conduct ablation experiments to prove the effectiveness of our model designs, including various message embedding strategies and loss combinations.
Subsequently, we evaluate the bit accuracy and fidelity of our method across different values of $\lambda_\text{recon}$, $\lambda_\text{msg}$, and $\lambda_\text{off}$ to analyze the sensitivity.
The results are evaluated on the Blender \cite{nerf} and LLFF \cite{llff} datasets with $N_L=32$ bits.

\noindent\tb{Various message embedding Strategies.}
We compare our SH-aware message embedding module with three strategies in \Cref{table:diff_variants}, including updating all the attributes: ``Offest$_{\text{all}}$'', the DC components of SH features: ``Offset$_{\text{dc}}$'', and the residuals of SH features: ``Offest$_{\text{rest}}$''.
As shown, Offest$_{\text{all}}$ achieves unsatisfactory reconstruction quality (row 1) as it alters the original 3D structure to embed messages during optimization.
Moreover, Offset$_{\text{dc}}$ (row 2) and Offest$_{\text{rest}}$ (row 3) are inferior to our SH-aware module (row 4), proving the correctness of our motivation.

\noindent\tb{Different Loss Combinations.}
We explore the optimal loss combinations in \Cref{table:diff_losses}.
Compared to the original 3DGS model (row 1), only using message loss $\mathcal{L}_\text{msg}$ (row 2) will significantly reduce the reconstruction quality.
Though simultaneously minimizing the message loss $\mathcal{L}_\text{msg}$ and reconstruction loss $\mathcal{L}_\text{recon}$ (row 3) can alleviate the decline in fidelity, it can only achieve sub-optimal results due to some large SH offsets.
Thus, we design the offset loss $\mathcal{L}_\text{off}$ to restrict the deviation of SH offsets, which achieves the optimal reconstruction quality (row 4).

\noindent\tb{Various Hyper-parameter Values.}
We analyze the sensitivity of 3 hyper-parameters: $\lambda_\text{recon}$, $\lambda_\text{msg}$, and $\lambda_\text{off}$ in \Cref{fig:sensitivity}.
For simplification, we only change the value of one hyper-parameter, while keeping the other two at their default values.
Visually, as increasing $\lambda_\text{recon}$ from 0 to 1, the PSNR rises with a subtle decline in bit accuracy.
When $\lambda_\text{msg}\in[0, 0.03]$, there is a significant change in performance. 
However, this effect gradually diminishes as $\lambda_\text{msg}\in[0.03, 0.1]$.
$\lambda_\text{off}=10$ is a watershed in performance, as values above and below it will influence the trade-off.
Thus, we choose $\lambda_\text{recon}=1$, $\lambda_\text{msg}=0.03$, and $\lambda_\text{off}=10$ for the optimal overall performance.

\begin{table}[!t]
\caption{\tb{Different loss combinations} with $N_L$=$32$ bits on Blender \cite{nerf} and LLFF \cite{llff} datasets. 
The first row denotes the original 3DGS, and $\mathcal{L}_\text{recon}=\mathcal{L}_\text{rgb}$ + $\mathcal{L}_\text{lpips}$ indicates the reconstruction loss.}
\centering
\setlength{\tabcolsep}{1.5mm}
\label{table:diff_losses}
\resizebox{0.88\linewidth}{!}{
\begin{tabular}{ccc|cccc}
\toprule
$\mathcal{L}_\text{msg}$ & $\mathcal{L}_\text{recon}$ & $\mathcal{L}_\text{off}$ & Bit Acc & PSNR     & SSIM  & LPIPS \\\midrule
&            &           & 53.41  & \textit{inf} & 1.0000 & 0.0000\\
\ding{52}      &            &           & 100.00 & 31.79    & 0.9604 & 0.0379\\
\ding{52}      & \ding{52}       &           & 99.26  & 36.88    & 0.9831 & 0.0101\\
\rowcolor{MyGray}
\ding{52}      & \ding{52}       & \ding{52}      & 99.04  & 39.40    & 0.9939 & 0.0022 \\\bottomrule
\end{tabular}}
\vspace{-1em}
\end{table}

\section{Conclusion}
In this paper, we present \textit{GuardSplat}, a novel watermarking framework to protect the copyright of 3DGS assets.
Specifically, we build an efficient message decoder via CLIP-guided Message Decoupling Optimization, enabling high-capacity and efficient 3DGS watermarking.
Moreover, we tailor a SH-aware message embedding module for 3DGS to seamlessly embed the messages while maintaining fidelity, meeting the demands for invisibility and security.
We further propose an anti-distortion message extraction module to achieve strong robustness against various visual distortions.
Experiments demonstrate that our \textit{GuardSplat} outperforms the baselines and achieves fast training speed.

\noindent\textbf{Acknowledgement.}
This project is partially supported by the NSFC (U22A2095), the Major Key Project of PCL under Grant PCL2024A06, and the Project of Guangdong Provincial Key Laboratory of Information Security Technology (Grant No. 2023B1212060026).


{
\small
\bibliographystyle{ieeenat_fullname}
\bibliography{main}
}

\clearpage
\renewcommand\thesection{\Alph{section}}
\renewcommand\thefigure{S\arabic{figure}}
\renewcommand{\thetable}{S\arabic{table}}
\setcounter{page}{1}
\setcounter{section}{0}
\setcounter{figure}{0}
\setcounter{table}{0}

\maketitlesupplementary

\setlength{\abovecaptionskip}{8pt}

\section{Overview}
In this supplementary material, we further provide more discussions, implementation details, and results as follows:
\begin{itemize}
\item \Cref{sec:Decoder} depicts the architecture of our message decoder guided by CLIP \cite{clip}, and we also conduct a comparison for watermarking efficiency against the state-of-the-art methods.
\item \Cref{sec:Security} conducts an additional evaluation for security, exploring whether the watermarks can be simply removed from model files.
\item \Cref{sec:Visual_comp} illustrates the visualization results of various ablations in \Cref{table:diff_variants,table:diff_losses,} of the main paper.
\item \Cref{sec:Results} reports more results, including the quantitative results on larger-capacity messages $N_L=\{64, 72\}$, bit accuracy across various rendering situations, and the zoomed-in rendering results between watermarked and original views.
\end{itemize}

\section{Decoder Architecture and Watermarking Speed}\label{sec:Decoder}
\begin{figure}[!t]
\includegraphics[width=0.47\textwidth]{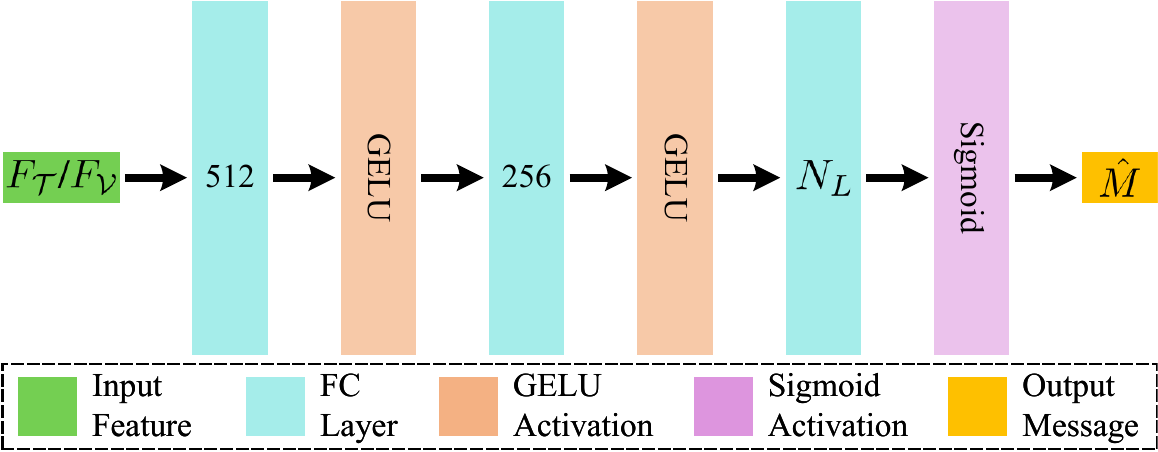}
\caption{\tb{The architecture of our message decoder.} Given an output feature $F_\mathcal{T}$ or $F_\mathcal{V}$, we first pass it through two FC layers with GELU activations, where their channels are set to 512 and 256, respectively.
Then, we map the feature to the binary message using a $N_L$-channel FC layer and a Sigmoid activation.
}
\label{fig:network}
\end{figure}

\begin{figure}[!t]
\centering
\includegraphics[width=0.47\textwidth]{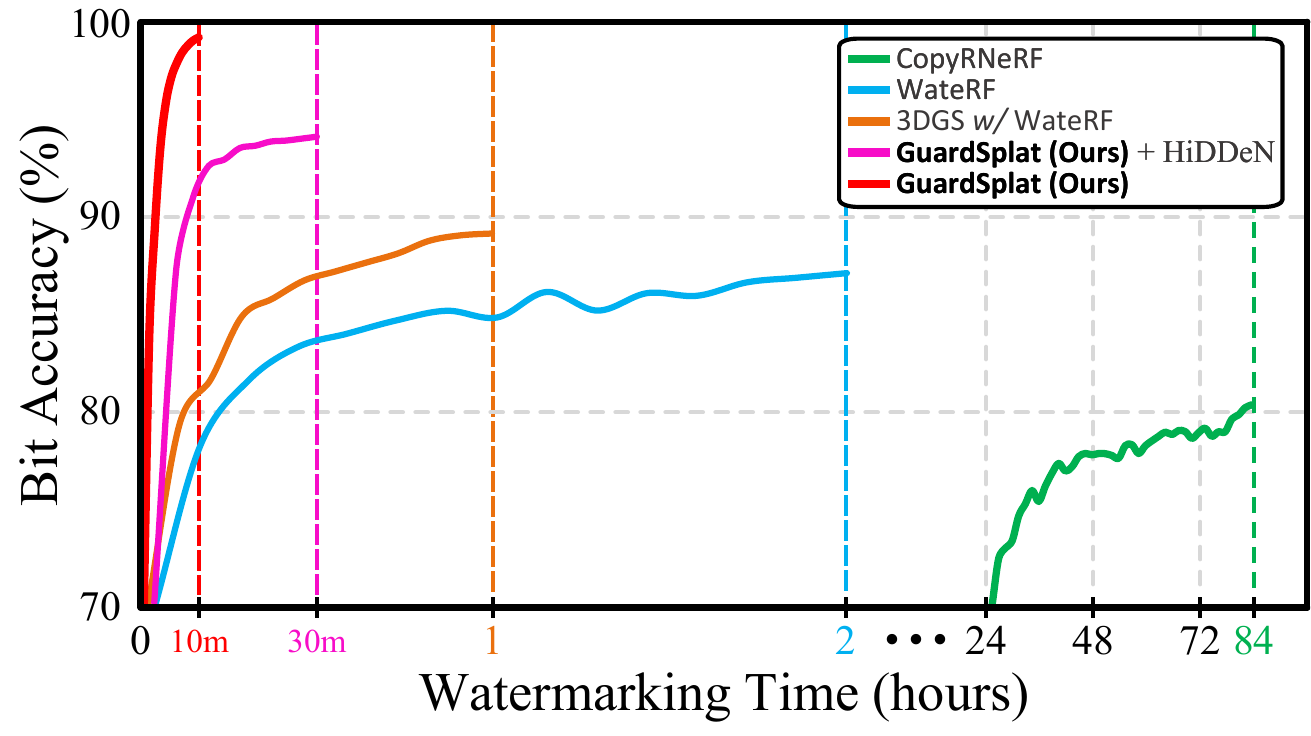}
\caption{\tb{Training accuracy curves} with $N_L=32$ bits on Blender \cite{nerf} dataset.
Our \tb{GuardSplat} achieves high training efficiency, which only takes 10 minutes to watermark a 3D asset.}
\label{fig:time}
\end{figure}

As shown in \cref{fig:network}, our message decoder only consists of 3 fully-connected (FC) layers, which can accurately map the CLIP textual features to the corresponding binary messages after a 5-minute optimization.
Thanks to CLIP's rich representation, our decoder can achieve excellent performance with minimal parameter size.
We also investigate the watermarking efficiency between our \textit{GuardSplat} and state-of-the-art methods.
As shown in the training accuracy curve in \Cref{fig:time}, our \textit{GuardSplat} achieves the highest efficiency, which only takes 10 minutes to watermark a pre-trained 3DGS asset.

\section{Additional Evaluation for Security}\label{sec:Security}
We conduct additional experiments to evaluate the security of our \textit{GuardSplat} in \Cref{table:additional}, investigating whether the malicious users can remove the watermarks from the model file by pruning the $K\%$ of Gaussians, where $K\in\{5,10,15,20,25\}$.
``Bottom $K$'' denotes pruning $K$ of low-opacity Gaussians, while ``random'' denotes randomly pruning $K$ of the Gaussians.
As demonstrated, our \textit{GuardSplat} still achieves a bit accuracy of 98.74\% when 25\% of the low-opacity Gaussians are removed, indicating that simply removing low-opacity Gaussians does not effectively attack our method.
Though randomly removing the Gaussians can lead to a significant decline in bit accuracy, it also greatly affects the reconstruction quality (\textit{i.e.,} PSNR, SSIM, and LPIPS), resulting in low-fidelity rendering.
This experimental result demonstrates that the malicious cannot directly remove the watermarks from the model file, verifying the security of our \textit{GuardSplat}.

\section{Additional Visual Comparisons} \label{sec:Visual_comp}
\subsection{Various Message Embedding Strategies}
\begin{figure}
\centering
\includegraphics[width=0.47\textwidth]{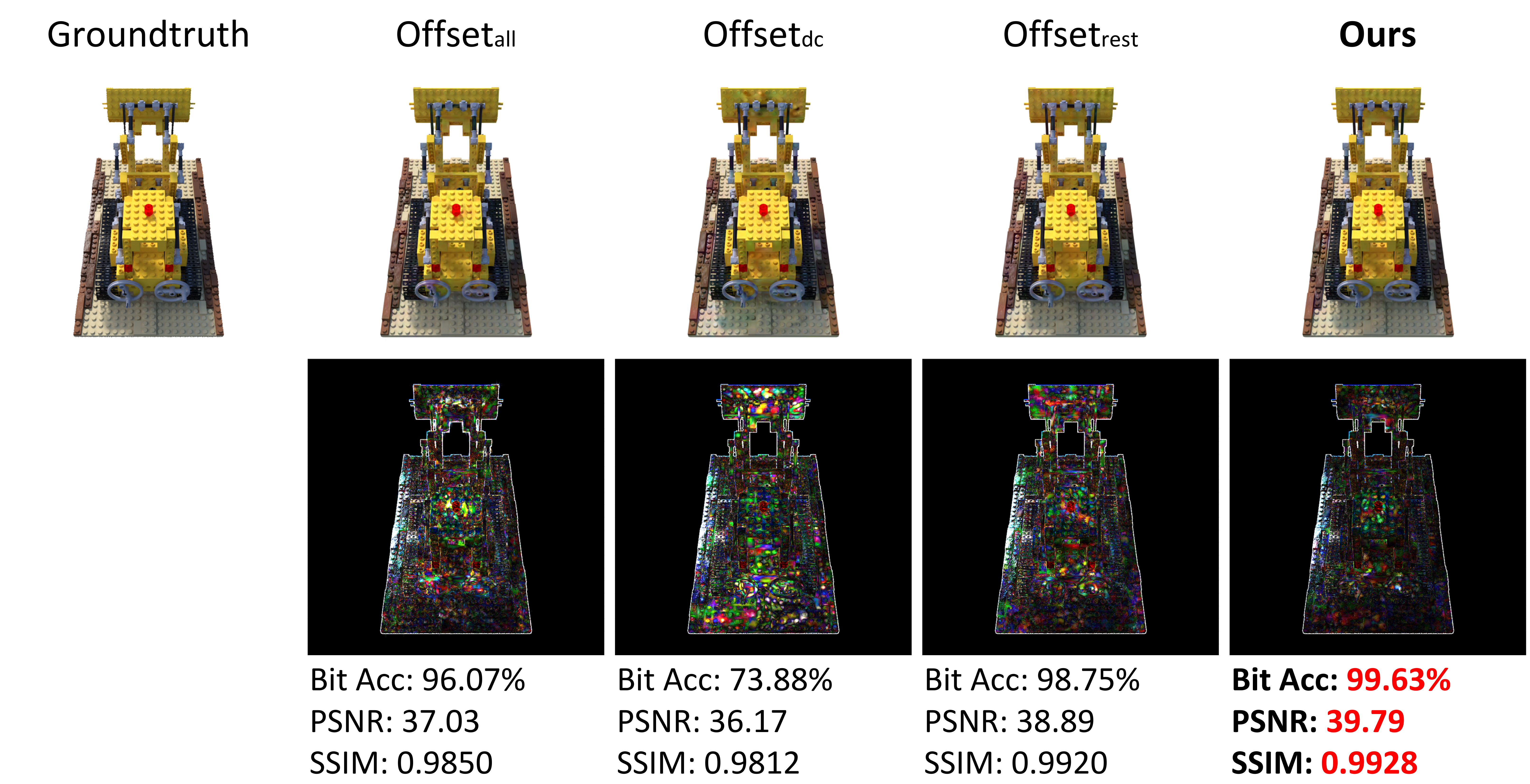}
\caption{\tb{Visual comparisons between various message embedding strategies and our SH-aware module.} Heatmaps at the bottom show the differences ($\times10$) between the watermarked and Groundtruth.
\textcolor{red}{Red} text indicates the best performance.}
\label{fig:watermark_embeddings}
\end{figure}

In the main paper, we explore the performance under various message embedding strategies with $N_L=32$ bits (see quantitative results in Table \textcolor{red}{3}).
For better comparisons, we further visualize the results of various message embedding strategies in \Cref{fig:watermark_embeddings}.
As shown, the proposed SH-aware module achieves superior bit accuracy and reconstruction quality to the competitors.

\subsection{Various Loss Combinations}
\begin{figure}
\centering
\includegraphics[width=0.47\textwidth]{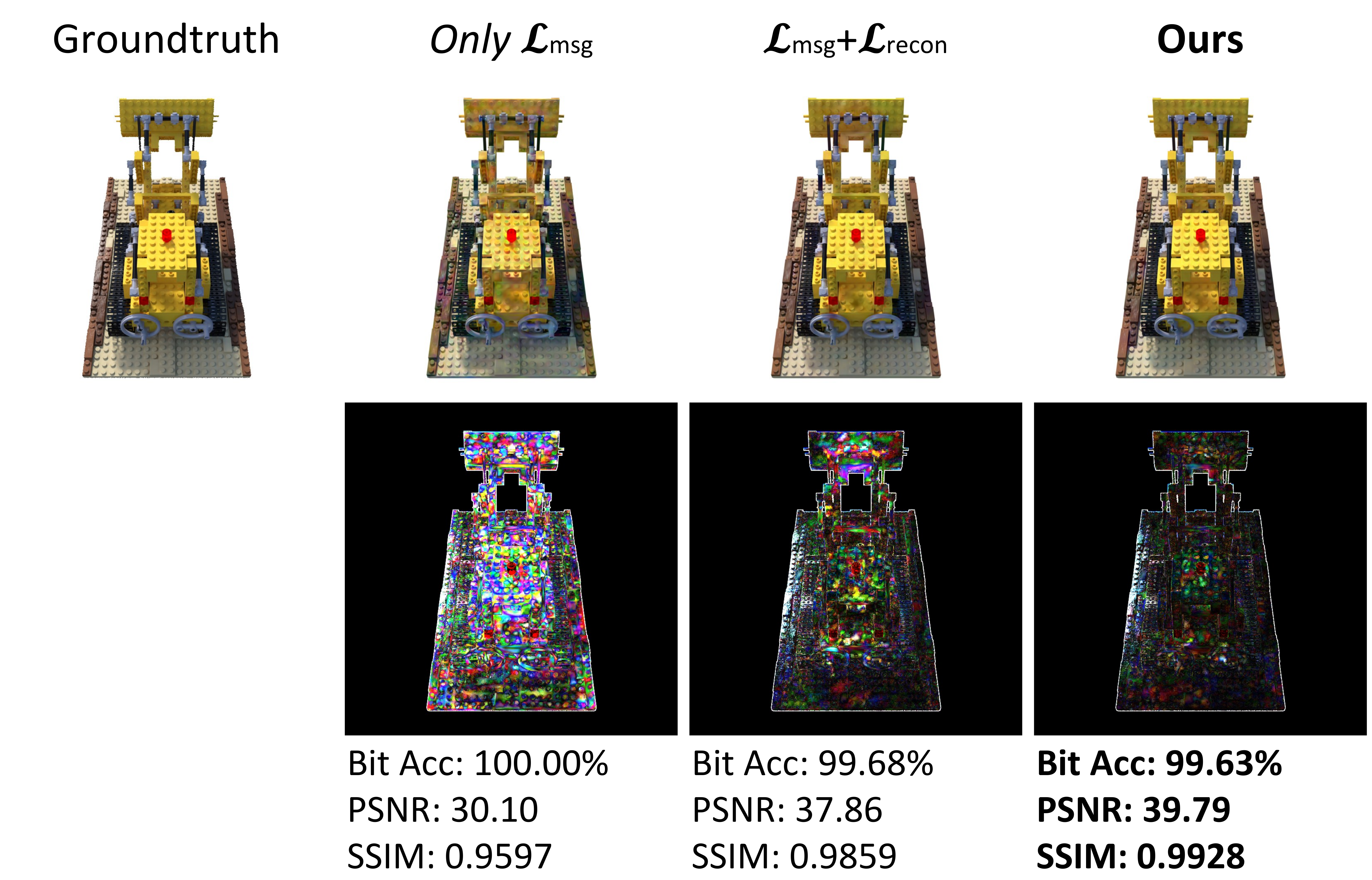}
\caption{\tb{Visual comparisons of various loss combinations.} ``Ours'' denotes the combination of $\mathcal{L}_\text{msg}+\mathcal{L}_\text{recon}+\mathcal{L}_\text{off}$.
Heatmaps at the bottom show the differences ($\times10$) between the watermarked and Groundtruth.
\tb{Bold} text indicates the best overall performance.
}
\label{fig:losses}
\end{figure}

In the main paper, we quantitatively compare the performance across various loss combinations in Table \textcolor{red}{4}.
We also conduct a visual comparison of these ablation variants in \Cref{fig:losses}.
As shown, ``$\mathcal{L}_\text{\text{recon}}+\mathcal{L}_\text{\text{msg}}+\mathcal{L}_\text{\text{off}}$'' achieves the best performance in bit accuracy and reconstruction quality.

\begin{table}[!t]
\caption{\tb{Security analysis across various pruning ratios} $\mathtt{K}\%$.
\texttt{Bottom K} denotes removing $\mathtt{K}\%$ of the low-opacity Gaussians, while \texttt{Random} denotes randomly removing $\mathtt{K}\%$ of the Gaussians.
}
\label{table:additional}
\setlength{\tabcolsep}{0.9mm}
\footnotesize
\centering
\begin{tabular}{cccccccccc}
\toprule
\multirow{3}{*}{\%} & \multicolumn{4}{c}{\texttt{Bottom $K$}} & & \multicolumn{4}{c}{\texttt{Random}} \\\cmidrule{2-5}\cmidrule{7-10}
&Bit Acc&PSNR &SSIM &LPIPS & &Bit Acc&PSNR &SSIM &LPIPS \\\midrule
5          &99.04 &39.38&0.9939&0.0022& &98.59 &37.76&0.9916&0.0033\\
10          &99.02 &39.06&0.9937&0.0025& &96.87 &36.35&0.9891&0.0047\\
15          &98.99 &38.68&0.9933&0.0031& &94.68 &35.14&0.9832&0.0063\\
20          &98.94 &38.33&0.9928&0.0037& &91.98 &33.98&0.9779&0.0081\\
25          &98.74 &37.87&0.9922&0.0041& &88.59 &31.50&0.9721&0.0103\\\bottomrule
\end{tabular}
\end{table}

\section{More Results}\label{sec:Results}
\subsection{Quantitative Results on Larger-Capacity Messages}
To further investigate the superiority of our \textit{GuardSplat} in capacity, we supplement the results on larger message lengths ($N_L\in\{64, 72\}$) in \Cref{table:larger}. As demonstrated, the bit accuracy and reconstruction quality of our 72-bit results are still higher than the state-of-the-art methods on $N_L\in\{16,32,48\}$ bits reported in the main paper (see Table \textcolor{red}{1}), significantly improve the capacity of existing baselines.

\subsection{Bit Accuracy across Various Rendering Situations}
\begin{figure}
\centering
\includegraphics[width=0.47\textwidth]{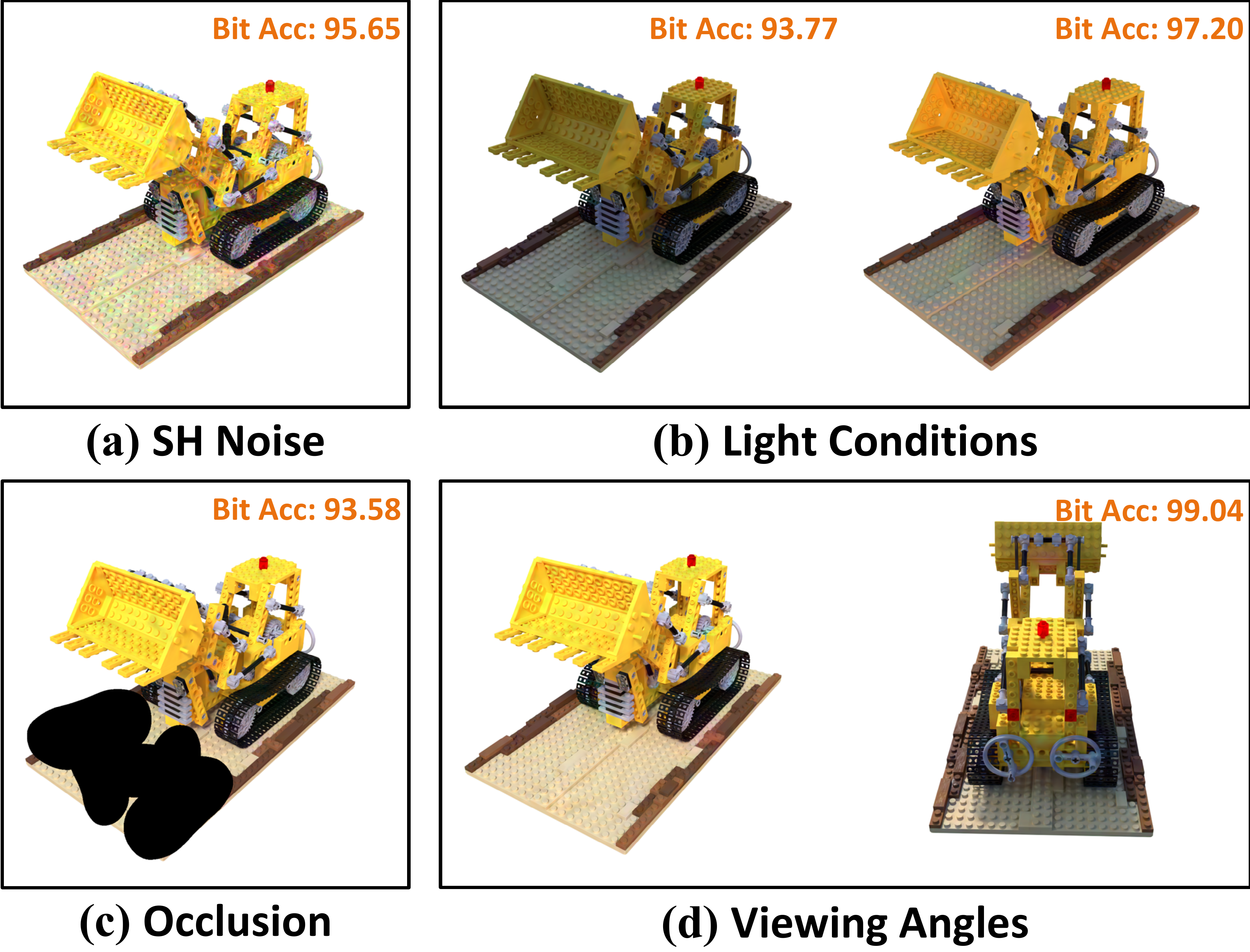}
\caption{Bit accuracy across various rendering parameters.}
\label{fig:SH}
\end{figure}

We explore the extraction accuracy of learned SH offsets across the following situations: \textbf{1)} SH Noise; \textbf{2)} Light Conditions; \textbf{3)} Occlusions; and \textbf{4)} Viewing Angles.
Specifically, to simulate different lighting conditions, we first train a 3DGS asset of ``Lego'' from the TensoIR \cite{TensoIR} dataset in "RGBA" mode. We then freeze all Gaussian attributes while optimizing the SH features to adapt to various illumination scenarios, such as ``light'', ``sunset'', and ``city''.
We train only the SH offsets in ``RGBA'' mode and add them to the SH features of other lighting modes for evaluation.
As shown in \Cref{fig:SH}, \textit{GuardSplat} achieves good robustness against SH noise \textbf{(a)} and light conditions \textbf{(b)} by adding noise to SH features in training.
Since the occluded areas can be removed by segmentation models (\textit{e.g.,} Segment Anything Model \cite{sam}, and Grounding DINO \cite{grounding_dino}), we train \textit{GuardSplat} to extract messages from randomly masked views ($\leq20\%$).
It improves the robustness of our \textit{GuardSplat} against various occlusions \textbf{(c)}.
\textit{GuardSplat} is inherently robust to various viewing angles \textbf{(d)} since it is designed for 3D.

\begin{table}[!t]
\caption{\tb{Quantitative results of our \textit{GuardSplat}} on Blender \cite{nerf} and LLFF \cite{llff} datasets with $N_L\in\{64, 72\}$ bits.}
\label{table:larger}
\centering
\begin{tabular}{c|cccc}
\toprule
$N_L$& Bit Acc & PSNR & SSIM  & LPIPS \\\midrule
64  & 97.41  & 37.76 & 0.9899 & 0.0040 \\
72  & 96.64  & 36.47 & 0.9866 & 0.0053 \\\bottomrule
\end{tabular}
\end{table}

\begin{figure}
\centering
\includegraphics[width=0.47\textwidth]{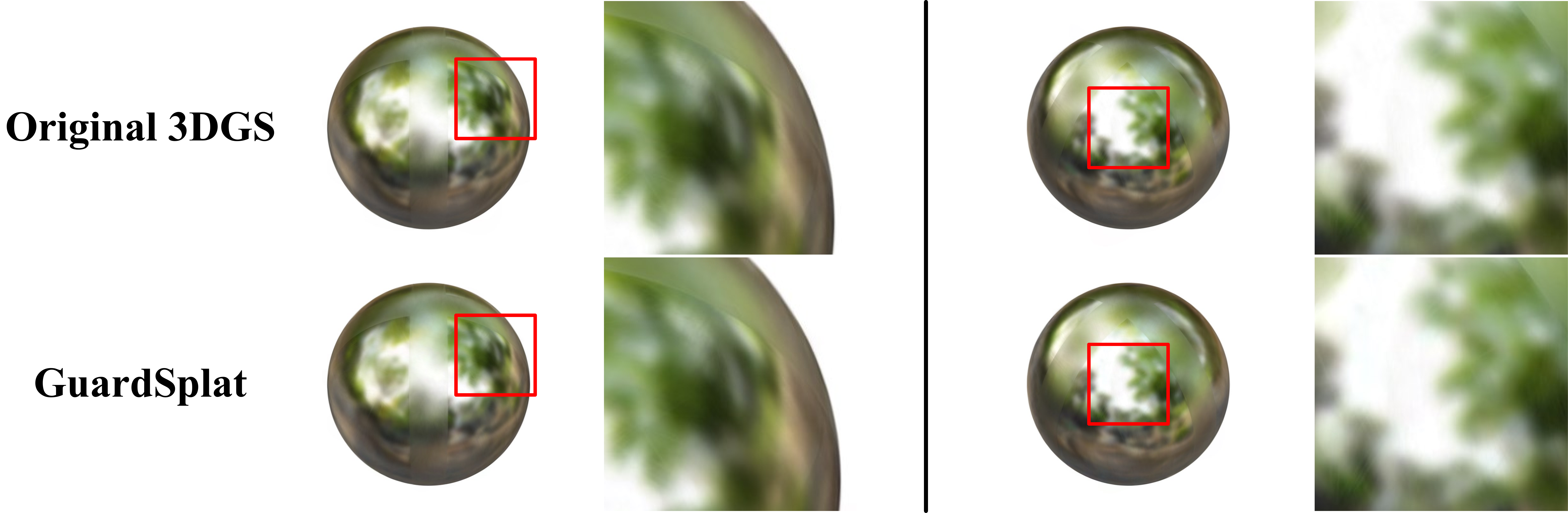}
\caption{Zoomed-in rendering results between the original 3DGS and our \textit{GuardSplat}.}
\label{fig:zoom}
\end{figure}

\subsection{Zoomed-in Rendering Results}
Since SH features produce highly realistic shading and shadowing, altering them may reduce fidelity, especially in the specular areas.
To clearly show how the SH offsets are changing the rendering results, we conduct a visual comparison of zoomed-in rendering results between the original 3DGS and our \textit{GuardSplat} of ``ball'' on the Shiny \cite{refnerf} dataset in \Cref{fig:zoom}.
As shown, \textit{GuardSplat} can preserve the original metallic luster of assets.

\end{document}